\documentclass[10pt,twocolumn,letterpaper]{article}

\usepackage{wacv}
\usepackage{times}
\usepackage{epsfig}
\usepackage{graphicx}
\usepackage{amsmath}
\usepackage{amssymb}
\usepackage[]{algorithm2e}

\wacvfinalcopy 

\ifwacvfinal
\def\assignedStartPage{9876} 
\fi

\ifwacvfinal
\usepackage[breaklinks=true,bookmarks=false]{hyperref}
\else
\usepackage[pagebackref=true,breaklinks=true,colorlinks,bookmarks=false]{hyperref}
\fi

\ifwacvfinal
\else
\fi

\begin{document}

\title{Rotated Ring, Radial and Depth Wise Separable Radial Convolutions}

\author{Wolfgang Fuhl\\
University of T\"ubingen\\
Sand 14, 72076 T\"ubingen, Germany\\
{\tt\small wolfgang.fuhl@uni-tuebingen.de}
\and
Enkelejda Kasneci\\
University of T\"ubingen\\
Sand 14, 72076 T\"ubingen, Germany\\
{\tt\small enkelejda.kasneci@uni-tuebingen.de}
}

\maketitle

\begin{abstract}
Simple image rotations significantly reduce the accuracy of deep neural networks. Moreover, training with all possible rotations increases the data set, which also increases the training duration. In this work, we address trainable rotation invariant convolutions as well as the construction of nets, since fully connected layers can only be rotation invariant with a one-dimensional input. On the one hand, we show that our approach is rotationally invariant for different models and on different public data sets. We also discuss the influence of purely rotational invariant features on accuracy. The rotationally adaptive convolution models presented in this work are more computationally intensive than normal convolution models. Therefore, we also present a depth wise separable approach with radial convolution. Link to CUDA code \url{https://atreus.informatik.uni-tuebingen.de/seafile/d/8e2ab8c3fdd444e1a135/}
\end{abstract}

\section{Introduction}

Deep neural networks have become the state of the art in image processing. Many applications requiring image classification~\cite{lecun1995convolutional,UMUAI2020FUHL}, landmark regression~\cite{wu2018group}, image segmentation~\cite{long2015fully,ICCVW2019FuhlW,CAIP2019FuhlW,ICCVW2018FuhlW} or 3D reconstruction~\cite{li2017deep}, validation~\cite{ICMV2019FuhlW,NNVALID2020FUHL}, gaze estimation~\cite{NNETRA2020}, object detection~\cite{WTCDAHKSE122016,WTCDOWE052017,WDTTWE062018,VECETRA2020,CORR2017FuhlW1,ETRA2018FuhlW}, shape estimation~\cite{WTDTWE092016,WTDTE022017,WTE032017}, saliency prediction~\cite{DWTE022017,AGAS2018} are realized by deep neural networks. This requires a large annotated data set~\cite{krizhevsky2012imagenet}, as well as additional data manipulation~\cite{madry2017towards}, to obtain robust and efficient networks. This data manipulation requires additional computational effort during training and extends the data set many times over. One of the most frequently used data manipulations is the mirroring and rotation of images. However, mirroring can only be applied to two axes in 2D images and is therefore inexpensive. In contrast, the rotation offers an infinite number of possibilities. Modern training methods use optimizers like SGD~\cite{bottou1991stochastic} or ADAM~\cite{kingma2014adam} which, in turn, use one or more gradient moments~\cite{qian1999momentum}. In addition, techniques such as Batch Normalization~\cite{ioffe2015batch} have been developed to stabilize the training and improve generalization. The training of really deep nets was only possible with Residual Layers~\cite{he2016deep} and there are many techniques to include data manipulation according to optimization strategies in the training~\cite{madry2017towards}. While all these techniques further stabilize the training and promote generalization, object rotation is still a major challenge together with the resource consumption reduction~\cite{AAAIFuhlW}. Examples of applications include: carried eye trackers with adjustable cameras like the Dikablis Professional, food~\cite{marin2019recipe1m}, objects in a kitchen~\cite{Damen2018EPICKITCHENS}, galaxies~\cite{abd2017automatic}, the underwater world from the perspective of a diver, plankton~\cite{lumini2019deep}, grabbing robot arms with a moving camera~\cite{pinto2016curious} and many more. In these tasks, it is desirable to use neural networks, which by definition are rotation invariant, to reduce the training time and ensure that the objects can be detected under all possible rotations. The various applications summarized above have already produced some scientific work in the field of rotational invariance. 

The first group of approaches addresses the integration of rotation into the training data. Here, different transformations are applied to the data and a separate mesh is trained for each transformation. In the end, either the mean value of all nets~\cite{ciregan2012multi,dieleman2015rotation,laptev2016ti} or the maximum result of a rotation map of all nets is used \cite{RotCNN,ICIP}. The second group focuses on learning about the actual transformations \cite{dai2017deformable,hinton2011transforming,jaderberg2015spatial}. Here nets are trained, which transform the data to a uniform representation. As in the first group, it is necessary to rotate the data during training. The final group of approaches tackles the complete integration of trainable rotated filters \cite{wacv18,1604,2003,cohen2016group,cohen2016steerable,worrall2017harmonic,marcos2017rotation,zhou2017oriented}, to which this work contributes. This means that the built-in layers are learned holistically with the net and can be trained with the back propagation algorithms. This final group is also referred to as steerable approaches It is not necessary to transform the data for training for this group.

In this work, we propose two novel approaches for integrated rotation invariant training of deep neural networks. Our work can be seen as an extension of the rotated weights approaches~\cite{wacv18,1604}, which reduce the performance of neural networks drastically. This is completed by rotating the filter per filter level outgoing of the center and selecting the maximum response per rotation ring. Since our approach increases the computational costs of deep neural networks, we also propose a radial convolution approach which only consists of a radial weight vector. In addition, we further reduce the computational complexity of our approaches with the technique of depth wise separable convolutions, proposed in MobileNet~\cite{howard2017mobilenets}. We tested our layers with different models and on different publicly available data sets. Our main contributions are as follows:

\begin{description}
	\setlength\itemsep{-0.5em}
	\item[1] Rotated ring convolutions.
	\item[2] Separable depth wise rotated convolutions.
	\item[3] Radial Convolutions.
	\item[4] Publicly available CUDA implementations for both.
\end{description}

\section{Related Work}
Since the state of the art is divided into three main groups, we have divided this section into three subsections. The first group deals with the separate learning of transformations by individual neural networks and the subsequent combination for classifications. The second group learns how to transform input data to a unified representation, allowing any net to be used subsequently. In the last group, the filters are rotated and fully integrated into the back propagation for training. In the following subsections, individual approaches are described in detail.

\subsection{Data Augmentation and Ensemble building}
In this group, transformations are applied to the data during training and trained with a variety of nets. This ensemble of nets increases the accuracy of the transformed data while also increasing the calculation time because multiple nets have to be trained. In \cite{ciregan2012multi}, for example, such an ensemble was trained were each net learned its own transformation. For the application, the results of the nets were averaged and, in case of a classification, the maximum was selected. In \cite{dieleman2015rotation}, galaxies were transformed and merged. This representation was used for training. The mesh learned to see the galaxy from different points of view in parallel, which made it more robust and accurate. A very similar approach was presented in \cite{laptev2016ti}. Here, Siamese networks with shared weights were used. Instead of merging the feature vectors of these networks, a max pooling was applied. The output of the max pooling was used to train a subsequent classifier. Another approach is exhibited in \cite{RotCNN} which internally uses an ensemble of oriented edge detectors. The output of these edge filters is projected on a two-dimensional rotation map using different meshes. The map is then used by a classifier to determine the class and rotation. \cite{ICIP} presents a similar approach because it uses a set of meshes to create a two-dimensional rotational map. However, compared to \cite{RotCNN}, wavlets are used instead of oriented edge filters.

\subsection{The learning of Transformations}
Unsupervised learning is the first approach when learning transformations. For this purpose, \cite{hinton2011transforming} used an autoencoder and let it learn the reverse transformation. The autoencoder was trained with as many transformations as possible on the data and the classifier was trained on the output, which received a uniform input from the autoencoder. A similar approach was presented in \cite{jaderberg2015spatial}. Here, a small network with weakly monitored learning was trained to transform the input data into a uniform output. This small net can then be integrated at any point in the net. In \cite{dai2017deformable}, we went one step further and trained the network to learn deformable convolutions and a deformable region of interest pooling. However, deformable convolutions and a deformable regions have to be trained for the different transformations. As in the first group, the data with applied transformations must be available in the training for learning transformations. This is a big disadvantage because it prolongs the training and there is no guarantee that all transformations in the training have been considered.

\begin{figure*}[h]
	\centering
	\includegraphics[width=0.99\textwidth]{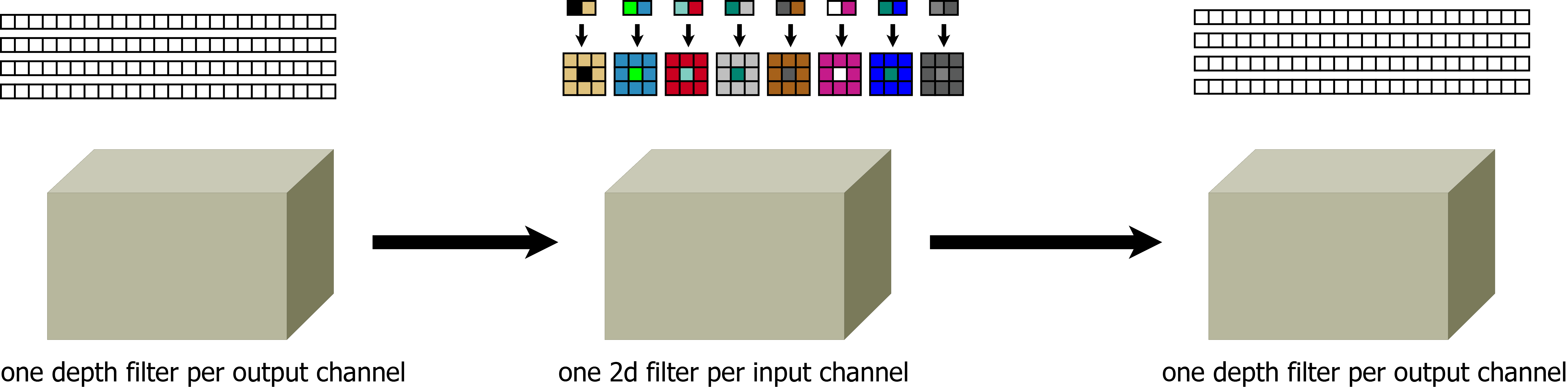}
	\caption{Visual description of the depth wise convolutions with the radial filter construction (RSDW). The left and ride side are the depth wise convolutions and in the center are the radial filters. In comparison to RAD the filters are applied as 2D Convolutions per channel and not as tensors.}
	\label{fig:RSDW}
\end{figure*}

\subsection{Steerable approaches}
This section explores the methods that transform filters and, as a result, do not require rotation of the training data. The network filters' ability to adapt, thus negating the need for a rotation of the input data, is the ultimate goal. These approaches are also called controllable approaches. The theoretical framework was presented in \cite{cohen2016steerable}. Based on the rotation of filters, a number of approaches have been proposed \cite{Wacv2018,1604,cohen2016group,cohen2016steerable,worrall2017harmonic,marcos2017rotation,zhou2017oriented} which differ only minimally. Informed by the work in \cite{cohen2016steerable}, group-equivariant convolution neural networks (GCNNs) were proposed in \cite{cohen2016group}. These consist of groups of rotated convolutions which are then used to perform a pooling operation along the output layers. The groups of convolutions are limited to 90 degrees. Harmonic nets (N-Nets) \cite{worrall2017harmonic} are an extension without the restriction of 90 degrees. N-Nets limit the filters to circular harmonics and require summing over several convolutions. That makes these nets, like many others, very computationally demanding when compared to standard convolutions. Another approach is the use of vector field nets \cite{marcos2017rotation,1604}. Here, the filter is rotated by predefined rotations and the maximum is selected from the output. In this approach, the maximum and the orientation are both passed, leading the naming of the nets. Oriented Response Networks (ORN) \cite{zhou2017oriented} are similar networks in that the number of orientations is arbitrary. For this purpose, the filter is rotated based on the orientations and the maximum is selected from the convolution. The difference between ORN's and vector field networks is that, in an ORN, the filter is actively rotated, but the amount of orientation is preliminarily fixed to four or eight. The response, as well as the orientation, is passed. Since this method is not rotation invariant, as is the case in vector fields, the orientation is passed to the next layer. The authors propose two ways to make ORN's rotation invariant. The first is a SIFT feature, such as alignment, and the second is maximum selection. A further variant of this rotating filters approach and a novel rotated pooling based module using the filter orientations was presented in Rotationally-Invariant Convolution Module (RIC) \cite{Wacv2018}. Here the authors have limited themselves to $3 \times 3$ convolutions and rotated only the weights between corners. This turns one filter into exactly four for all rotations. The other weights (corners and center) are identical. As in \cite{Wacv2018}, we do not use any rotation in the training phase and evaluate different rotations in the validation phase.




\section{Method}
\begin{figure}[h]
	\centering
	\includegraphics[width=0.45\textwidth]{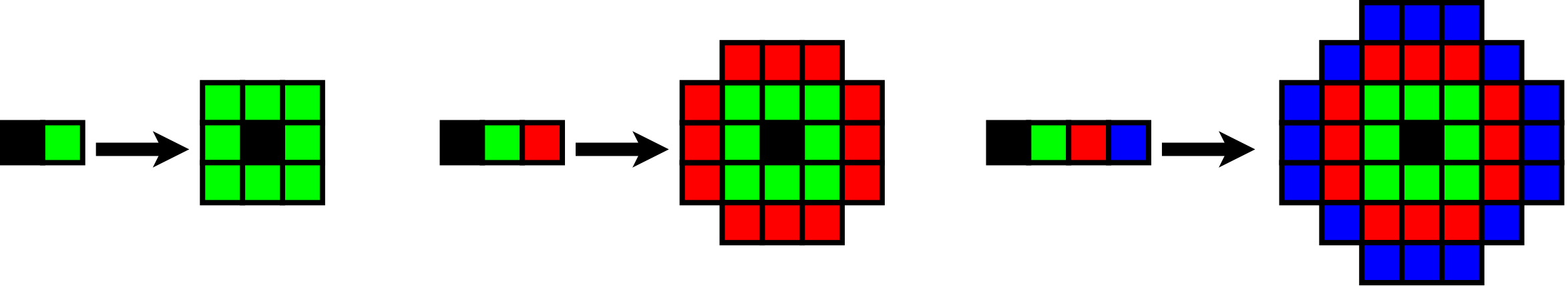}
	\caption{Visual description of the filter construction of our radial convolutions (RAD). The same color stands for the same value. On the left side of the arrow are the parameters without bias term and on the right side, the applied convolution. Each input convolution tensor has other radial parameters per input channel.}
	\label{fig:RAD}
\end{figure}
Figure~\ref{fig:RAD} shows the radial convolutions. As you can see, the weights to be learned (always on the left side of an arrow) correspond to an integer distance from the center of the filter (always on the right side of an arrow). The filters are then constructed so that the distance of the index in the filter to the center is the index of the parameters.

\begin{equation}
	F_{c_{in},c_{out}}(i,j) = W_{c_{in},c_{out}}(round(sqrt(i^2 + j^2)))
\label{eq:RAD}
\end{equation}

This indexing is described in Equation~\ref{eq:RAD} where $i,j$ is the 2D Filter Index and $W$ is the weight field. Since the convolutions are tensors and not only 2D filters, each channel in the tensor has its own weight field $c_{in}$. This is, of course, also valid for the output channels $c_{out}$ where each channel has a separate tensor. In the following, this method will be called RAD.

We have also extended this method with the technique of depth wise convolutions. In the first step, a tensor based depth wise convolution is performed, which determines the number of channels in the middle layer. This can be seen in Figure~\ref{fig:RSDW}. In the middle layer, the radial convolutions are applied not as tensor, but as individual 2D convolutions for each input channel. Thus, the first depth wise convolution determines the number of output and input channels in the middle layer in Figure~\ref{fig:RSDW}. At the end, a depth wise convolution is performed again to create a desired output depth. This method is called RSDW in the following.

\begin{equation}
F_{c_{in/out}}(i,j) = W_{c_{in/out}}(round(sqrt(i^2 + j^2)))
\label{eq:RSDW}
\end{equation}

Equation~\ref{eq:RSDW} shows the change for the middle section of Figure~\ref{fig:RSDW}. Since we only use one 2D convolution for each input channel and the input and output layers have the same size, we only need one additional filter index for the channels $ c_{in/out}$.

Both methods (RAD, RSDW) are by definition rotation invariant because each feature in e.g. a $3 \times 3$ field can be rotated around the central pixel without changing the multiplication equation with the filter. In addition, all pixels with the same distance from the center can be permuted arbitrarily without changing the multiplication equation. Of course, this is also a disadvantage because the learned features are less meaningful. Since this is generally already the case for purely rotation invariant features~\cite{wacv18,1604,2003,cohen2016group,cohen2016steerable,worrall2017harmonic,marcos2017rotation,zhou2017oriented}, however, it is not a major disadvantage.

\begin{figure*}[h]
	\centering
	\includegraphics[width=0.9\textwidth]{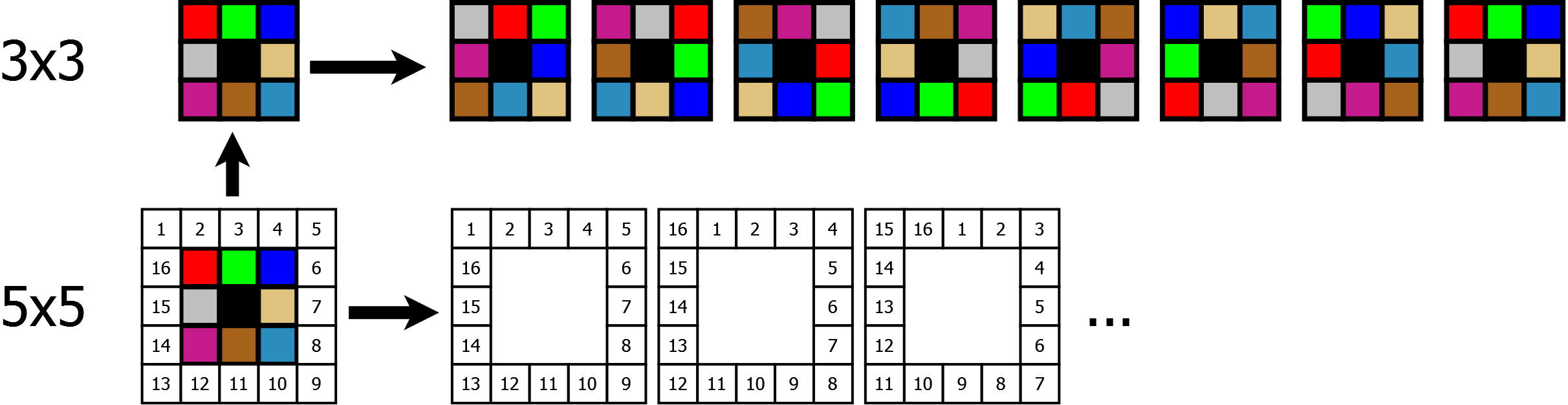}
	\caption{Visual description of the rotated ring convolutions (RING). In the top row the rotations for a $3 \times 3$ are shown. For a $5 \times 5$ convolution, the outer ring is rotated independently. This can be seen in the bottom row.}
	\label{fig:RING}
\end{figure*}

\begin{figure}[h]
	\centering
	\includegraphics[width=0.45\textwidth]{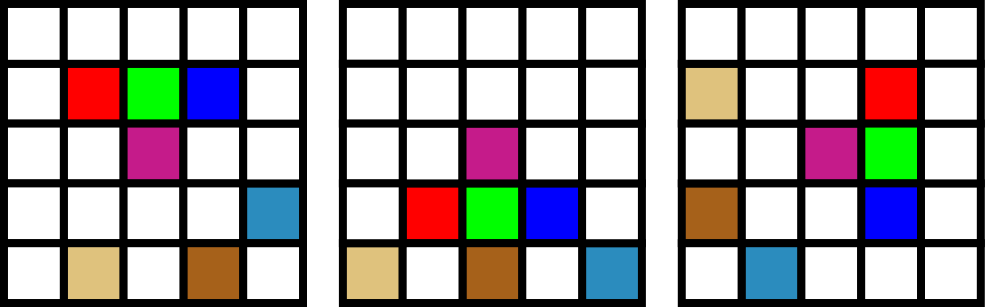}
	\caption{Visual example of features that rotate within themselves. All three patches represent the same feature and the two outer rings are always independently rotated around the central pixel.}
	\label{fig:RINGEX}
\end{figure}

With our third approach (RING), we address the reduced meaningfulness of the features from the first approach. For this purpose, we use a similar approach to the rotated filters~\cite{zhou2017oriented}. However, in our method, each individual ring of the filter rotates. This can be seen in Figure~\ref{fig:RING}. In the case of a $3 \times $3 convolution, this corresponds to the rotated filters of \cite{zhou2017oriented}. In the case of a $5 \times 5$ convolution (lower part of figure 3), another ring is added. This second or third ring, if the central weight is considered a separate ring, will also rotate independently. This independent rotation of the rings makes it possible to learn more than just the features with rotated filters. An example is shown in Figure~\ref{fig:RINGEX} where each pattern represents the same feature.

As you can see, the outer and inner ring always rotate, independently, around the central pixel. Another advantage of looking at each ring independently is that not all combinations of both rotations result in their own filters. Each ring gets its own convolution filters, thereby reducing the number convolutions to $8+16$ instead of $8*16$. At the end, the maximum of each ring is selected and added together with the bias term. Compared to the rotated filters~\cite{zhou2017oriented}, our method does not require a fixed number $F$ of filters which has to be set in advance because each ring is always completely rotated. Our method results in $\sum_{i=2}^{f} 2*(i+i-2)$ convolutions for symmetric filters with width and height $f$.

To train these rotating filters using the backpropagation algorithm, the error as well as the gradient must be rotated. To make this easier and to use the fast cuDNN convolutions, we have simply added several storage stages, which allow us to assign the error to each rotated filter and, subsequently, to each weight. This is similar to the idea behind rotated filters~\cite{zhou2017oriented}, the difference being that we add separate stages for multiple rings.

\begin{algorithm}
	\SetKwFunction{FMain}{Main}
	\SetKwProg{Fn}{Function}{ is}{end}
	\KwData{Data,Weights,WRot}
	\KwResult{Output,ORot}
	\Fn{Forward}{
		WRot=RotateRings(Weights)\;
		ORot=cuDNNFWD(WRot,Data)\;
		Output=Comp\&MaxPerRing(ORot)\;
	}
	\KwData{ErrorIn,ORot,WRot}
	\KwResult{ErrorOut,ERot}
	\Fn{Backward}{
		ERot=MaxPerRingBWD(ErrorIn,ORot)\;
		ErrorOut=cuDNNBWD(WRot,ERot)\;
	}
	\SetKwFunction{FMain}{Main}
	\SetKwProg{Fn}{Function}{is}{end}
	\KwData{ERot,Data}
	\KwResult{Grad}
	\Fn{CompGrad}{
		GRot=cuDNNCompGrad(Data,ERot)\;
		Grad=RotateRingsGrad(GRot)\;
	}
	\caption{Algorithmic description of the modifications to the convolution procedure for the forward, backward, and gradient computation work flow.}
	\label{alg:FWDBWDGRAD}
\end{algorithm}

Algorithm~\ref{alg:FWDBWDGRAD} describes modifications to the back propagation algorithm that were made in order to use the normal convolution together with the cuDNN implementations for convolution. In the forward algorithm, the filters have to be created based on the ring rotations (RotateRings). Afterwards, the cuDNN convolution can be used. To combine the results of the individual rings and to achieve the correct output depth, the combination and maximum selection must be executed after the convolution (Comp\&MaxPerRing). For the back propagation of the error, the input gradient must be assigned to the maxima selected in the forward step (MaxPerRingBWD). Following this, the cuDNN back propagation can be used. To calculate the gradients, the assigned error and the input data can be used directly with the cuDNN gradient calculation. Finally, only the calculated gradients have to be assigned to individual weights in the non-rotated filter (RotateRingsGrad). Later, the weights can be updated with any optimizer.

\begin{table}[h]
	\centering
	\caption{The necessary weights without bias term. Conv represents a normal tensor convolution. In case of RSDW, out1 stands for the output channels of the first depth wise convolution and out2 for the output channels of the last depth wise convolution.}
	\label{tbl:param}
	\begin{tabular}{lc}
		\textbf{Method} & \textbf{Weights} \\ \hline
		Conv $3 \times 3$ & 3*3*in*out \\
		Conv $5 \times 5$ & 5*5*in*out \\
		RIC $3 \times 3$ ~\cite{wacv18} & 5*in*out \\
		ORN $3 \times 3$ ~\cite{zhou2017oriented}& 3*3*in*out \\
		ORN $5 \times 5$ ~\cite{zhou2017oriented}& 5*5*in*out \\
		RAD $2$ (ours) & 2*in*out \\
		RAD $3$ (ours) & 3*in*out \\
		RSDW $2$ (ours) & (in*out1) + (2*out1) + (out1*out2) \\
		RSDW $3$ (ours) & (in*out1) + (3*out1) + (out1*out2) \\
		RING $3 \times 3$ (ours) & 3*3*in*out \\
		RING $5 \times 5$ (ours) & 5*5*in*out   
	\end{tabular}
\end{table}

Table~\ref{tbl:param} shows the required parameters for each approach. As you can see, no approach requires more parameters than a typical convolution. However, the number of parameters for RIC~\cite{wacv18}, as well as for RAD, is much lower. In the case of RSDW, it depends on the number of depth wise convolutions, where the depth of the first is $out1$ and the depth of the second is $out2$.

\begin{table}[h]
	\centering
	\caption{The complexity of one layer execution without biasterm. Conv represents a normal tensor convolution. In case of RSDW, out1 stands for the output channels of the first depth wise convolution and out2 for the output channels of the last depth wise convolution.}
	\label{tbl:complex}
	\begin{tabular}{lc}
		\textbf{Method} & \textbf{Complexity} \\ \hline
		Conv $3 \times 3$ & w*h*3*3*in*out \\
		Conv $5 \times 5$ & w*h*5*5*in*out \\
		RIC $3 \times 3$ ~\cite{wacv18}& 4*w*h*3*3*in*out \\
		ORN-F $3 \times 3$ ~\cite{zhou2017oriented}& F*w*h*3*3*in*out \\
		ORN-F $5 \times 5$ ~\cite{zhou2017oriented}& F*w*h*5*5*in*out \\
		RAD $2$ (ours) & w*h*3*3*in*out \\
		RAD $3$ (ours) & w*h*5*5*in*out \\
		RSDW $2$ (ours) & (w*h*out1) * ((in) + (3*3) + (out2)) \\
		RSDW $3$ (ours) & (w*h*out1) * ((in) + (5*5) + (out2)) \\
		RING $3 \times 3$ (ours) & w*h*3*3*in*out \\
		RING $5 \times 5$ (ours) & (8*w*h*in*out)*(3*3 + 2*5*5)
	\end{tabular}
\end{table}

In Table~\ref{tbl:complex}, the complexity of a layer is provided based on the input depth $in$ and surface $w \times h$. As you can see, only the depth wise convolutions for certain depths can reduce complexity. Once constant factor in all other methods is the expense, which is greater than ordinary convolutions. In the case of ORN~\cite{zhou2017oriented}, this is the predefined number of rotated filters $F$. For RIC~\cite{wacv18}, it is the four integrated rotations and, for RING, the sum of the rotations of the individual rings ($\sum_{i=2}^{f} 2*(i+i-2)$).

\section{Neural Network Models}
\begin{figure*}[h]
	\centering
	\includegraphics[width=0.85\textwidth]{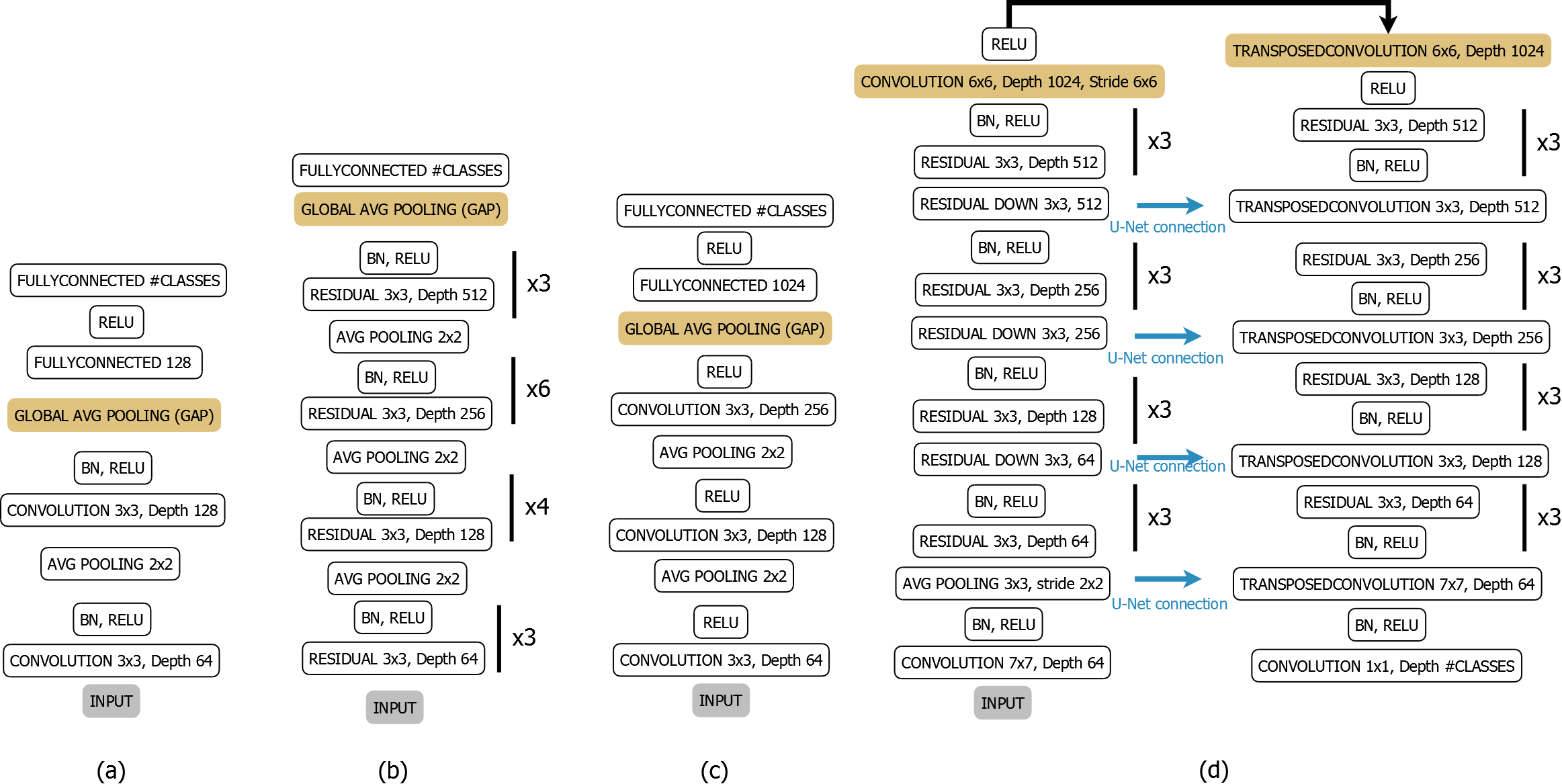}
	\caption{All used architectures in our evaluation. (a) is a small neural network model with batch normalization. (b) represents a ResNet-34 with batch normalization and residual blocks~\cite{he2016deep}. (c) is a small classical neural network model without batch normalization. (d) is a so called U-Net~\cite{ronneberger2015u} with residual blocks~\cite{he2016deep} and fully convolutional~\cite{long2015fully}.}
	\label{fig:models}
\end{figure*}

Figure~\ref{fig:models} shows the architectures that we used in our experiments. To showcase the applicability of the rotation invariant layers, we used diverging architectures. As an example, the first model (Figure~\ref{fig:models} a)) is a small model with batch normalization and the second model (Figure~\ref{fig:models} b)) a ResNet-34. The first model was developed to evaluate the applicability of our layers to batch normalization. The second model was developed to show that the layers can also be used with modern residual networks. In addition, we used a small classical neural network (Figure~\ref{fig:models} c)). We used this to show the impact of different rotation invariant layers on models without batch normalization. The last model (Figure~\ref{fig:models} d), is a fully convolutional neural network~\cite{long2015fully} with additional connections between resolution stages. Those models are called U-Nets~\cite{ronneberger2015u} and the interconnections improve the semantic segmentation result. This network was only used with the VOC2012~\cite{pascalvoc2012} data set and the semantic segmentation task. For training and evaluation, we used the DLIB~\cite{king2009dlib} library for deep neural networks. In the library, we have also integrated our rotation invariant layers and the state of the art approaches by which we compare our proposed approach.

As exhibited in all models, the last layer is always a pooling stage which reduces the output tensor dimension to one. This is to achieve the model's rotation invariance since the tensor itself also contains spatial information for the following fully connected stages. We will show the difference empirically in our first evaluation.

\section{Data sets}
In this section,  all the data sets, training parameters and algorithms for weight initialization utilized in our work are described. In addition, we provide the parameters and optimization procedures as well as the batch size used per data set. We used as little data augmentation as possible to ensure an easy reproduction of our results and described this in detail too.

\textbf{CIFAR10}~\cite{krizhevsky2009learning} is a publicly available data set with a total of 60,000 images. It has ten classes which have to be estimated for a given image. Each image in this data set has a resolution of $32 \times 32$ and three color channels (red, green, and blue). The training set consists of 50,000 images with 5,000 images per class. For validation, 10,000 images are provided with 1,000 images per class.

\textit{\textbf{Training:} As optimizer, we used ADAM~\cite{kingma2014adam} with a weight deacay of $5*10^{-5}$, momentum one with $0.9$ and momentum two with $0.999$. The batch size was fixed to fifty during training. The initial learning rate was set to $10^{-3}$ and the training was conducted for 300 epochs. After each 50 epochs, the learning rate was reduced by $10^{-1}$. The weights of our models were initialized using formula 16 from \cite{glorot2010understanding} and all bias terms were set to 0 initially. We did not use any data augmentation during training. For preprocessing, we used constant mean subtraction (mean-red $122.782$, mean-green $117.001$, mean-blue $104.298$) and division by $256.0$ for the input image. For the evaluation, we evaluated the rotations $0^{\circ}$, $90^{\circ}$, $180^{\circ}$ and $270^{\circ}$ separately to show the rotation invariance of the proposed approaches. It has to be noted that no rotation was used during training.}

\textbf{CIFAR100}~\cite{krizhevsky2009learning} is also a public data set, but with one hundred instead of ten classes. It has the same task as CIFAR10, which is selecting a class based on a given image. The image resolution for CIFAR100 is $32 \times 32$ with three color channels (red, green, and blue). The data set contains 60,000 images and is split into a training and a validation set. The training set contains 50,000 images with 500 examples per class and the validation set contains 10,000 images with 100 examples per class. Therefore, CIFAR100 is a balanced data set with the same number of images as in CIFAR10.

\textit{\textbf{Training:} As an optimizer we used SGD~\cite{bottou1991stochastic} with first momentum $0.9$. Weight deacay was set to $5*10^{-4}$ and we used a fixed batch size of 50 during the training. The initial learning rate was set to $10^{-1}$ and we trained for a total of 300 epochs. After each round of 50 epochs, we reduced the learning rate by $10^{-1}$. We did not use any data augmentation, but, instead, used image preprocessing. This preprocessing was a constant mean subtraction (mean-red $122.782$, mean-green $117.001$, mean-blue $104.298$) and division by $256.0$ for the input image. For weight initialization, we used formula 16 from \cite{glorot2010understanding} and all bias terms were set to 0. For the evaluation, we evaluated the rotations $0^{\circ}$, $90^{\circ}$, $180^{\circ}$, and $270^{\circ}$ separately to show the rotation invariance of the proposed approaches. It has to be noted that no rotation was used during training.}

\textbf{VOC2012}~\cite{pascalvoc2012} is a publicly available data set. It contains annotations for object detection, classification and semantic segmentation. We used the annotations for semantic segmentation only. In semantic segmentation, the task is to classify each pixel in an image regarding its class affiliation. Overall, the VOC2012 data set has twenty classes and the background class. Each image can contain multiple objects, but not all twenty objects have to be present. It is also possible that the same object class is present multiple times. The training set consists of 3,507 segmented objects on 1,464 images and the validation set has 3,422 segmented objects on 1,449 images. Additionally, there is a third image set without any annotations. This set can be used for initializing weights using unsupervised training similar to an autoencoder. In our training and evaluation, we did not use the third set. In addition, the data set is unbalanced, making it even more difficult to use.

\textit{\textbf{Training:}  We used a fixed batch size of ten during training. The initial learning rate was set to $10^{-1}$. The entire training procedure contained 800 epochs, whereby after each round of 200 epochs the learning rate was reduced by $10^{-1}$. As an optimizer we used SGD with momentum~\cite{qian1999momentum} set to $0.9$ and a weight decay of $1*10^{-4}$. For weight initialization, we used formula 16 from \cite{glorot2010understanding} and all bias terms were set to 0. For data augmentation, we used a cropping of $227 \times 227$ regions out of each input image. Additionally, we used random color offsets during training. As preprocessing of the images, we used constant mean subtraction (mean-red $122.782$, mean-green $117.001$, mean-blue $104.298$) and divided each value by $256.0$. For the evaluation, we evaluated the rotations $0^{\circ}$, $90^{\circ}$, $180^{\circ}$, and $270^{\circ}$ separately to show the rotation invariance of the proposed approaches. It has to be noted that no rotation was used during training.}

\section{Evaluation}

In general, it should be noted here that rotationally invariant features always perform worse when compared to rotationally sensitive features if evaluated for only one rotation. This is due to the fact that, for some tasks, rotation sensitive features contain more general information~\cite{Wacv2018,zhou2017oriented,marcos2017rotation,1604}. However, on average across all rotations, rotationally invariant traits performed more effectively. Since our models were trained without additional data manipulation and rotation, we limited the evaluation to four simple rotations, ($0^{\circ}, 90^{\circ}, 180^{\circ}, and 270^{\circ}$), which can be calculated without interpolation. In all tables, we share the results for each rotation to show that the trained features are rotation invariant when compared to conventional convolutions.

For comparison, we consistently replaced all convolution layers with rotationally invariant layers, with the exception of rotational pooling (RP)~\cite{Wacv2018}. Here, the authors have shown that the process works best if only the first layer is used. Therefore, in our evaluation, we only used the rotational pooling in the first layer together with the rotational invariant layer when $\& RP$ was specified. For the RIC~\cite{Wacv2018} approach, we only used filters with a size $3 \times 3$, since no construction rule was specified for larger filters. For ORN~\cite{zhou2017oriented}, we used all rotations every time, where the number was specified with $-XY$ each. To make the ORN~\cite{zhou2017oriented} approach rotationally invariant, we used the ORPooling~\cite{zhou2017oriented} outlined by the authors.

\begin{table}
	\centering
	\setlength\tabcolsep{0.05em}
	\caption{Shows the accuracy on CIFAR10 with the global pooling removed. RP refers to the rotational pooling of \cite{Wacv2018}. Only the first convolution is replaced by the rotational invariant filters and the rotational pooling is applied subsequently. For all evaluations we used model c) from Figure~\ref{fig:models}.}
	\label{tbl:archi}
	\begin{tabular}{lcccc}
		\textbf{Method} & \textbf{$0^{\circ}$} & \textbf{$90^{\circ}$} & \textbf{$180^{\circ}$} & \textbf{$270^{\circ}$} \\ \hline
		CNN $3 \times 3$ & 79.71\% & 28.62\% & 33.65\% & 29.04\% \\
		CNN $5 \times 5$~\cite{wacv18} & \textbf{81.36\%}  & 30.25\% & 36.08\% &  29.42\% \\
		RIC $3 \times 3$~\cite{wacv18} & 69.74\%  & 29.98\% & 31.11\% &  28.59\% \\
		RIC \& RP $3 \times 3$~\cite{wacv18} & 53.21\%  & 53.21\% & 53.21\% &  53.21\% \\
		ORN-8 $3 \times 3$~\cite{zhou2017oriented} & 72.89\%  & 34.66\% & 34.36\% &   33.65\% \\
		ORN-16 $5 \times 5$~\cite{zhou2017oriented} & 76.31\% & 36.21\% & 37.59\% & 36.41\%\\
		ORN-8 \& RP $3 \times 3$~\cite{zhou2017oriented} & 58.17\% & 58.17\% & 58.17\% & 58.17\%\\
		ORN-16 \& RP $5 \times 5$~\cite{zhou2017oriented} & 61.48\% & 61.48\% & 61.48\% & 61.48\%\\
		RAD $2$ (ours) & 65.05\% & 29.14\% & 30.99\% & 29.85\% \\
		RAD $3$ (ours) & 68.92\% & 25.61\% & 34.68\% & 32.76\% \\
		RAD $2$ \& RP  (ours) & 50.26\% & 50.26\% & 50.26\% & 50.26\% \\
		RAD $3$ \& RP  (ours) & 57.35\% & 57.35\% & 57.35\% & 57.35\% \\
		RSDW $2$ (ours) & 62.21\%  & 26.43\% & 29.87\% & 27.39\% \\
		RSDW $3$ (ours) & 66.71\% & 26.41\% & 31.35\% & 29.18\% \\
		RSDW $2$ \& RP (ours) & 48.01\% & 48.01\% & 48.01\% & 48.01\% \\
		RSDW $3$ \& RP (ours) & 55.76\% & 55.76\% & 55.76\% & 55.76\% \\
		RING $3 \times 3$ (ours) & 72.89\%  & 34.66\% & 34.36\% & 33.65\% \\
		RING $5 \times 5$ (ours) & 79.58\% & 37.93\% & 39.90\% & 38.59\% \\
		RING \& RP $3 \times 3$ (ours) & 58.17\% & 58.17\% & 58.17\% & 58.17\%  \\
		RING \& RP $5 \times 5$ (ours) & 62.90\% & \textbf{62.90\%} & \textbf{62.90\%} & \textbf{62.90\%}  
	\end{tabular}
\end{table}

Table~\ref{tbl:archi} shows the results on CIFAR10 using the model c) from Figure~\ref{fig:models}. Global pooling has been removed for this evaluation. As you can see, when comparing the entries without rotational pooling~\cite{Wacv2018}, the invariant features are not sufficient to make the entire network invariant to a rotation on the input data. One approach to make the entire network invariant is to use rotational pooling~\cite{Wacv2018}. Here, for each different rotation of the filter, the output is rotated as well. In the case of back propagation, the error is also rotated. This makes the nets rotationally invariant as seen in Table~\ref{tbl:archi} for the entries with RP. We have used RP together with the invariant features only in the first layer because the authors in \cite{Wacv2018} have shown that is where it works the best. If you compare the results with RP from Table~\ref{tbl:archi} and the results with global pooling from Table~\ref{tbl:cif10}, you can see that all nets together with the rotation invariant features are rotation invariant. Global pooling also allows for high accuracy. Consequently, we will use global pooling for the following evaluations instead of RP. Another observation that stands out in the tables \ref{tbl:archi}, \ref{tbl:cif10} and \ref{tbl:cif100} is that RING $3 \times 3$ and ORN-8 $3 \times 3$~\cite{zhou2017oriented} always achieve the same results. This is because the RING method only rotates one ring and there are eight filter rotations of ORN-8~\cite{zhou2017oriented}. Thus, for the input $3 \times 3$, both approaches are identical in our evaluation. For ORN~\cite{zhou2017oriented}, we used all rotations of the outermost filter ring. For ORN~\cite{zhou2017oriented}, the complete filter was always rotated, making a difference in the filter size of $5 \times 5$.

\begin{table}
	\centering
	\setlength\tabcolsep{0.05em}
	\caption{Shows the accuracy on CIFAR10 where the letters a), b), and c) denote the models in Figure~\ref{fig:models}. Where possible, each method was evaluated with different filter sizes $X \times X$.}
	\label{tbl:cif10}
	\begin{tabular}{lcccc}
		\textbf{Method} & \textbf{$0^{\circ}$} & \textbf{$90^{\circ}$} & \textbf{$180^{\circ}$} & \textbf{$270^{\circ}$} \\ \hline
		CNN $3 \times 3$(a)) & 77.85\% & 35.84\% & 47.38\% & 35.61\%   \\
		CNN $5 \times 5$(a)) & \textbf{78.20\%}  & 35.20\% & 43.80\% &  35.96\%  \\
		RIC $3 \times 3$(a))~\cite{wacv18} & 66.73\% & 66.73\% & 66.73\% & 66.73\%   \\
		ORN-8 $3 \times 3$(a))~\cite{zhou2017oriented} & 74.88\% & 74.88\% & 74.88\% & 74.88\%  \\
		ORN-16 $5 \times 5$(ab))~\cite{zhou2017oriented} & 75.15\% & 75.15\% & 75.15\% & 75.15\%  \\
		RAD $2$(ours) (a)) & 64.41\% & 64.41\% & 64.41\% & 64.41\%   \\
		RAD $3$(ours) (a)) & 67.17\% & 67.17\% & 67.17\% & 67.17\%   \\
		RSDW $2$(ours) (a)) & 60.56\% & 60.56\% & 60.56\% & 60.56\%   \\
		RSDW $3 $(ours) (a)) & 62.75\% & 62.75\% & 62.75\% & 62.75\%   \\
		RING $3 \times 3$(ours) (a)) & 74.88\% & 74.88\% & 74.88\% & 74.88\%   \\
		RING $5 \times 5$(ours) (a)) & 76.10\% & \textbf{76.10\%} & \textbf{76.10\%} & \textbf{76.10\%}   \\ \hline
		CNN $3 \times 3$(b)) & 81.93\% & 30.61\% & 34.82\% & 31.79\%   \\
		CNN $5 \times 5$(b)) & \textbf{87.78\%} & 33.56\% & 37.34\% & 30.39\%   \\
		RIC $3 \times 3$(b))~\cite{wacv18} & 70.74\% & 70.74\% & 70.74\% & 70.74\%  \\
		ORN-8 $3 \times 3$(b))~\cite{zhou2017oriented} & 77.63\% & 77.63\% & 77.63\% & 77.63\%   \\
		ORN-16 $5 \times 5$(b))~\cite{zhou2017oriented} & 79.03\% & 79.03\% & 79.03\% & 79.03\%   \\
		RAD $2$(ours) (b)) & 67.06\% & 67.06\% & 67.06\% & 67.06\%  \\
		RAD $3$(ours) (b)) & 69.83\% & 69.83\% & 69.83\% & 69.83\%   \\
		RSDW $2$(ours) (b)) & 64.45\% & 64.45\% & 64.45\% & 64.45\%   \\
		RSDW $3 $(ours) (b)) & 65.82\% & 65.82\% & 65.82\% & 65.82\%   \\
		RING $3 \times 3$(ours) (b)) & 77.63\% & 77.63\% & 77.63\% & 77.63\%  \\
		RING $5 \times 5$(ours) (b)) & 79.96\% & \textbf{79.96\%} & \textbf{79.96\%} & \textbf{79.96\%}  \\ \hline
		CNN $3 \times 3$(c)) & 81.01\% & 31.61\% & 39.12\% & 32.06\% \\
		CNN $5 \times 5$(c)) & \textbf{84.36\%}  & 32.25\% & 38.08\% &  26.42\%  \\
		RIC $3 \times 3$(c))~\cite{wacv18} & 66.31\% & 66.31\% & 66.31\% & 66.31\%  \\
		ORN-8 $3 \times 3$(c))~\cite{zhou2017oriented} & 74.27\% & 74.27\% & 74.27\% & 74.27\%   \\
		ORN-16 $5 \times 5$(c))~\cite{zhou2017oriented} & 75.68\% & 75.68\% & 75.68\% & 75.68\%  \\
		RAD $2$(ours) (c)) & 65.70\% & 65.70\% & 65.70\% & 65.70\%   \\
		RAD $3$(ours) (c)) & 66.84\% & 66.84\% & 66.84\% & 66.84\%   \\
		RSDW $2$(ours) (c)) & 61.40\% & 61.40\% & 61.40\% & 61.40\%   \\
		RSDW $3 $(ours) (c)) & 64.25\% & 64.25\% & 64.25\% & 64.25\%   \\
		RING $3 \times 3$(ours) (c)) & 74.27\% & 74.27\% & 74.27\% & 74.27\%   \\
		RING $5 \times 5$(ours) (c)) & 76.83\% & \textbf{76.83\%} & \textbf{76.83\%} & \textbf{76.83\%}   
	\end{tabular}
\end{table}

\begin{table}
	\centering
	\setlength\tabcolsep{0.05em}
	\caption{Shows the accuracy on CIFAR100 where the letters a), b), and c) denote the models in Figure~\ref{fig:models}. Where possible, each method was evaluated with different filter sizes $X \times X$.}
	\label{tbl:cif100}
	\begin{tabular}{lcccc}
		\textbf{Method} & \textbf{$0^{\circ}$} & \textbf{$90^{\circ}$} & \textbf{$180^{\circ}$} & \textbf{$270^{\circ}$} \\ \hline
		CNN $3 \times 3$(a)) & 48.97\% & 23.30\% & 30.75\% & 24.05\%  \\
		CNN $5 \times 5$(a)) & \textbf{49.83\%} & 24.41\% & 32.99\% & 25.50\% \\
		RIC $3 \times 3$(a))~\cite{wacv18} & 39.10\% & 39.10\% & 39.10\% & 39.10\%  \\
		ORN-8 $3 \times 3$(a))~\cite{zhou2017oriented} & 47.41\% & 47.41\% & 47.41\% & 47.41\% \\
		ORN-16 $5 \times 5$(a))~\cite{zhou2017oriented} & 48.50\% & 48.50\% & 48.50\% & 48.50\%  \\
		RAD $2$(ours) (a)) & 37.66\% & 37.66\% & 37.66\% & 37.66\%  \\
		RAD $3$(ours) (a)) & 39.51\% & 39.51\% & 39.51\% & 39.51\% \\
		RSDW $2$(ours) (a)) & 33.56\% & 33.56\% & 33.56\% & 33.56\%  \\
		RSDW $3 $(ours) (a)) & 34.20\% & 34.20\% & 34.20\% & 34.20\%  \\
		RING $3 \times 3$(ours) (a))& 47.41\% & 47.41\% & 47.41\% & 47.41\%  \\
		RING $5 \times 5$(ours) (a))& 49.03\% & \textbf{49.03\%} & \textbf{49.03\%} & \textbf{49.03\%}  \\ \hline
		CNN $3 \times 3$(b)) & 67.31\% & 20.29\% & 21.16\% & 19.92\%  \\
		CNN $5 \times 5$(b)) & \textbf{71.20\%} & 19.39\% & 22.73\% & 18.90\% \\
		RIC $3 \times 3$(b))~\cite{wacv18} & 49.12\% & 49.12\% & 49.12\% & 49.12\%  \\
		ORN-8 $3 \times 3$(b))~\cite{zhou2017oriented} & 57.43\% & 57.43\% & 57.43\% & 57.43\%  \\
		ORN-16 $5 \times 5$(b))~\cite{zhou2017oriented} & 60.39\% & 60.39\% & 60.39\% & 60.39\% \\
		RAD $2$(ours) (b)) & 45.20\% & 45.20\% & 45.20\% & 45.20\%  \\
		RAD $3$(ours) (b)) & 48.33\% & 48.33\% & 48.33\% & 48.33\% \\
		RSDW $2$(ours) (b)) & 34.82\% & 34.82\% & 34.82\% & 34.82\%  \\
		RSDW $3 $(ours) (b)) & 36.91\% & 36.91\% & 36.91\% & 36.91\% \\
		RING $3 \times 3$(ours) (b)) & 57.43\% & 57.43\% & 57.43\% & 57.43\%  \\
		RING $5 \times 5$(ours) (b)) & 62.40\% & \textbf{62.40\%} & \textbf{62.40\%} & \textbf{62.40\%} \\ \hline
		CNN $3 \times 3$(c))& 48.53\% & 18.37\% & 22.29\% & 18.14\% \\
		CNN $5 \times 5$(c)) & \textbf{50.30\%} & 19.81\% & 25.01\% & 18.92\% \\
		RIC $3 \times 3$(c))~\cite{wacv18} & 36.01\% & 36.01\% & 36.01\% & 36.01\%  \\
		ORN-8 $3 \times 3$(c))~\cite{zhou2017oriented} & 43.86\% & 43.86\% & 43.86\% & 43.86\%  \\
		ORN-16 $5 \times 5$(c))~\cite{zhou2017oriented} & 45.72\% & 45.72\% & 45.72\% & 45.72\%   \\
		RAD $2$(ours) (c)) & 35.06\% & 35.06\% & 35.06\% & 35.06\%   \\
		RAD $3$(ours) (c)) & 37.95\% & 37.95\% & 37.95\% & 37.95\% \\
		RSDW $2$(ours) (c)) & 30.19\% & 30.19\% & 30.19\% & 30.19\%  \\
		RSDW $3$(ours) (c)) & 32.73\% & 32.73\% & 32.73\% & 32.73\% \\
		RING $3 \times 3$(ours) (c)) & 43.86\% & 43.86\% & 43.86\% & 43.86\%  \\
		RING $5 \times 5$(ours) (c)) & 46.89\% & \textbf{46.89\%} & \textbf{46.89\%} & \textbf{46.89\%} 
	\end{tabular}
\end{table}

In Tables \ref{tbl:cif10} and \ref{tbl:cif100}, the results can be observed on CIFAR10 and CIFAR100 for the models a), b), and c) with global pooling. All rotationally invariant layers and approaches show the rotational invariance clearly in comparison to ordinary convolutions because they offer the same results for each rotation. The least expensive and least parameterized method in our evaluation is RSDW, which uses the approach of depth wise convolutions~\cite{} together with radial convolutions (RAD). RSDW also has the worst results of all rotation invariants, but still maintains a better record, on average, than conventional convolutions when compared to all rotations. The RAD approach significantly improves the results and has both lower calculation costs and parameters compared to RIC~\cite{Wacv2018}. On the other hand, RIC, with a radius of 2 for model b), is always significantly better than RAD, which had a radius of 3 for model b). If RIC~\cite{Wacv2018} is compared directly with RAD, the calculation costs for RIC~\cite{Wacv2018} are four times higher and the number of parameters, in the case of RAD with a radius of 2, are almost three times higher.

Likewise, it can be seen in Tables \ref{tbl:cif10} and \ref{tbl:cif100} that, for the orientation $0^\circ$, the CNN with a filter size of $5 \times 5$ always performs best. This is also the orientation of the training data. For all other orientations, RING with a filter size of $5 \times 5$ is best. The second best is ORN-16 with a filter size of $5 \times 5$. ORN-16 only needs 2/3 of the run time when compared to RING.

\begin{table}
	\centering
	\setlength\tabcolsep{0.2em}
	\caption{Shows the pixel wise accuracy on VOC2012. The model for all evaluations is d) from Figure~\ref{fig:models}.}
	\label{tbl:segi}
	\begin{tabular}{lcccc}
		\textbf{Method} & \textbf{$0^{\circ}$} & \textbf{$90^{\circ}$} & \textbf{$180^{\circ}$} & \textbf{$270^{\circ}$} \\ \hline
		CNN $3 \times 3$(d)) & \textbf{84.65\%} & 62.85\% & 73.66\% & 53.78\%  \\
		RAD $3$(ours) (d)) & 71.25\% & 71.28\% & 71.81\% & 72.02\%  \\
		RING $3 \times 3$(ours) (d)) & 79.93\% & \textbf{79.16\%} & \textbf{79.85\%} & \textbf{80.12\%}
	\end{tabular}
\end{table}

As can be observed in Table~\ref{tbl:segi}, the rotation invariant filters also work for semantic segmentation. Note that we also scaled the validation data to $227 \times 227$ to achieve a one-dimensional vector in the central part of the network. As you can see, the resource-saving method RAD is already better than the conventional convolution while the more expensive RING method improves results significantly.

\section{Conclusion}
In this work, we have shown several new approaches for rotationally invariant feature extraction in deep neural networks. We compared our approaches with the state of the art on CIFAR10 and CIFAR100. Different models were trained and no rotation was applied to the training data. As the evaluation with four rotations reveals, our approaches are also rotation invariant. Futhermore, we tested our approaches in fully convolutional neural networks for semantic segmentation on the VOC2012 data set. In addition to the description and evaluation, we provide the CUDA implementations of our approaches as well as a description of how they can be integrated into existing frameworks. Future work will focus on head mounted eye trackers~\cite{WF042019}, since eye cameras can be shifted and rotated. Through this effort, we would like to further show the advantages of rotation invariant approaches also for scan path classification~\cite{C2019,FFAO2019} as well as eye movement detection~\cite{FCDGR2020FUHL,fuhl2018simarxiv,ICMIW2019FuhlW1,ICMIW2019FuhlW2,EPIC2018FuhlW,FCDGR2020FUHL}.  In particular, we will look at the transfer between training on synthetic data and the use of genuine human data. In addition, rotation invariant features could improve saliency prediction as well~\cite{DWTE022017,AGAS2018}.


{\small
\bibliographystyle{ieeefullname}
\bibliography{egbib}

\begin{thebibliography}{10}\itemsep=-1pt

\bibitem{abd2017automatic}
Mohamed Abd El~Aziz, IM Selim, and Shengwu Xiong.
\newblock Automatic detection of galaxy type from datasets of galaxies image
  based on image retrieval approach.
\newblock {\em Scientific Reports}, 7(1):1--9, 2017.

\bibitem{2003}
Vincent Andrearczyk, Julien Fageot, Valentin Oreiller, Xavier Montet, and
  Adrien Depeursinge.
\newblock Local rotation invariance in 3d cnns.
\newblock {\em arXiv preprint arXiv:2003.08890}, 2020.

\bibitem{bottou1991stochastic}
L{\'e}on Bottou.
\newblock Stochastic gradient learning in neural networks.
\newblock {\em Proceedings of Neuro-N{\i}mes}, 91(8):12, 1991.

\bibitem{ciregan2012multi}
Dan Ciregan, Ueli Meier, and J{\"u}rgen Schmidhuber.
\newblock Multi-column deep neural networks for image classification.
\newblock In {\em 2012 IEEE conference on computer vision and pattern
  recognition}, pages 3642--3649. IEEE, 2012.

\bibitem{cohen2016group}
Taco Cohen and Max Welling.
\newblock Group equivariant convolutional networks.
\newblock In {\em International conference on machine learning}, pages
  2990--2999, 2016.

\bibitem{cohen2016steerable}
Taco~S Cohen and Max Welling.
\newblock Steerable cnns.
\newblock {\em arXiv preprint arXiv:1612.08498}, 2016.

\bibitem{dai2017deformable}
Jifeng Dai, Haozhi Qi, Yuwen Xiong, Yi Li, Guodong Zhang, Han Hu, and Yichen
  Wei.
\newblock Deformable convolutional networks.
\newblock In {\em Proceedings of the IEEE international conference on computer
  vision}, pages 764--773, 2017.

\bibitem{Damen2018EPICKITCHENS}
Dima Damen, Hazel Doughty, Giovanni~Maria Farinella, Sanja Fidler, Antonino
  Furnari, Evangelos Kazakos, Davide Moltisanti, Jonathan Munro, Toby Perrett,
  Will Price, and Michael Wray.
\newblock Scaling egocentric vision: The epic-kitchens dataset.
\newblock In {\em European Conference on Computer Vision (ECCV)}, 2018.

\bibitem{dieleman2015rotation}
Sander Dieleman, Kyle~W Willett, and Joni Dambre.
\newblock Rotation-invariant convolutional neural networks for galaxy
  morphology prediction.
\newblock {\em Monthly notices of the royal astronomical society},
  450(2):1441--1459, 2015.

\bibitem{pascalvoc2012}
M. Everingham, L. Van~Gool, C.~K.~I. Williams, J. Winn, and A. Zisserman.
\newblock The {PASCAL} {V}isual {O}bject {C}lasses {C}hallenge 2012 {(VOC2012)}
  {R}esults.
\newblock
  http://www.pascal-network.org/challenges/VOC/voc2012/workshop/index.html.

\bibitem{wacv18}
Patrick Follmann and Tobias Bottger.
\newblock A rotationally-invariant convolution module by feature map
  back-rotation.
\newblock In {\em 2018 IEEE Winter Conference on Applications of Computer
  Vision (WACV)}, pages 784--792. IEEE, 2018.

\bibitem{Wacv2018}
Patrick Follmann and Tobias Bottger.
\newblock A rotationally-invariant convolution module by feature map
  back-rotation.
\newblock In {\em 2018 IEEE Winter Conference on Applications of Computer
  Vision (WACV)}, pages 784--792. IEEE, 2018.

\bibitem{WF042019}
W. Fuhl.
\newblock {\em Image-based extraction of eye features for robust eye tracking}.
\newblock PhD thesis, University of Tübingen, 04 2019.

\bibitem{UMUAI2020FUHL}
Wolfgang Fuhl.
\newblock From perception to action using observed actions to learn gestures.
\newblock {\em User Modeling and User-Adapted Interaction}, pages 1--18, 08
  2020.

\bibitem{C2019}
Wolfgang Fuhl, Efe Bozkir, Benedikt Hosp, Nora Castner, David Geisler, Thiago~C
  Santini, and Enkelejda Kasneci.
\newblock Encodji: encoding gaze data into emoji space for an amusing scanpath
  classification approach.
\newblock In {\em Proceedings of the 11th ACM Symposium on Eye Tracking
  Research \& Applications}, pages 1--4, 2019.

\bibitem{ICMIW2019FuhlW1}
W. Fuhl, N. Castner, and E. Kasneci.
\newblock Histogram of oriented velocities for eye movement detection.
\newblock In {\em International Conference on Multimodal Interaction Workshops,
  ICMIW}, 2018.

\bibitem{ICMIW2019FuhlW2}
W. Fuhl, N. Castner, and E. Kasneci.
\newblock Rule based learning for eye movement type detection.
\newblock In {\em International Conference on Multimodal Interaction Workshops,
  ICMIW}, 2018.

\bibitem{FFAO2019}
W. Fuhl, N. Castner, T.~C. Kübler, A. Lotz, W. Rosenstiel, and E. Kasneci.
\newblock Ferns for area of interest free scanpath classification.
\newblock In {\em Proceedings of the 2019 ACM Symposium on Eye Tracking
  Research \& Applications (ETRA)}, 06 2019.

\bibitem{ICCVW2018FuhlW}
W. Fuhl, N. Castner, L. Zhuang, M. Holzer, W. Rosenstiel, and E. Kasneci.
\newblock Mam: Transfer learning for fully automatic video annotation and
  specialized detector creation.
\newblock In {\em International Conference on Computer Vision Workshops,
  ICCVW}, 2018.

\bibitem{ETRA2018FuhlW}
W. Fuhl, S. Eivazi, B. Hosp, A. Eivazi, W. Rosenstiel, and E. Kasneci.
\newblock Bore: Boosted-oriented edge optimization for robust, real time remote
  pupil center detection.
\newblock In {\em Eye Tracking Research and Applications, ETRA}, 2018.

\bibitem{NNETRA2020}
W. Fuhl, H. Gao, and E. Kasneci.
\newblock Neural networks for optical vector and eye ball parameter estimation.
\newblock In {\em ACM Symposium on Eye Tracking Research \& Applications, ETRA
  2020}. ACM, 01 2020.

\bibitem{VECETRA2020}
W. Fuhl, H. Gao, and E. Kasneci.
\newblock Tiny convolution, decision tree, and binary neuronal networks for
  robust and real time pupil outline estimation.
\newblock In {\em ACM Symposium on Eye Tracking Research \& Applications, ETRA
  2020}. ACM, 01 2020.

\bibitem{ICCVW2019FuhlW}
W. Fuhl, D. Geisler, W. Rosenstiel, and E. Kasneci.
\newblock The applicability of cycle gans for pupil and eyelid segmentation,
  data generation and image refinement.
\newblock In {\em International Conference on Computer Vision Workshops,
  ICCVW}, 11 2019.

\bibitem{WDTTWE062018}
W. Fuhl, D. Geisler, T. Santini, T. Appel, W. Rosenstiel, and E. Kasneci.
\newblock Cbf:circular binary features for robust and real-time pupil center
  detection.
\newblock In {\em ACM Symposium on Eye Tracking Research \& Applications}, 06
  2018.

\bibitem{EPIC2018FuhlW}
W. Fuhl and E. Kasneci.
\newblock Eye movement velocity and gaze data generator for evaluation,
  robustness testing and assess of eye tracking software and visualization
  tools.
\newblock In {\em Poster at Egocentric Perception, Interaction and Computing,
  EPIC}, 2018.

\bibitem{ICMV2019FuhlW}
W. Fuhl and E. Kasneci.
\newblock Learning to validate the quality of detected landmarks.
\newblock In {\em International Conference on Machine Vision, ICMV}, 11 2019.

\bibitem{AAAIFuhlW}
W. Fuhl, G. Kasneci, W. Rosenstiel, and E. Kasneci.
\newblock Training decision trees as replacement for convolution layers.
\newblock In {\em Conference on Artificial Intelligence, AAAI}, 02 2020.

\bibitem{WTCDOWE052017}
W. Fuhl, T.~C. Kübler, D. Hospach, O. Bringmann, W. Rosenstiel, and E.
  Kasneci.
\newblock Ways of improving the precision of eye tracking data: Controlling the
  influence of dirt and dust on pupil detection.
\newblock {\em Journal of Eye Movement Research}, 10(3), 05 2017.

\bibitem{AGAS2018}
Wolfgang Fuhl, Thomas~C K{\"u}bler, Thiago Santini, and Enkelejda Kasneci.
\newblock Automatic generation of saliency-based areas of interest for the
  visualization and analysis of eye-tracking data.
\newblock In {\em VMV}, pages 47--54, 2018.

\bibitem{FCDGR2020FUHL}
Wolfgang Fuhl, Yao Rong, and Kasneci Enkelejda.
\newblock Fully convolutional neural networks for raw eye tracking data
  segmentation, generation, and reconstruction.
\newblock In {\em Proceedings of the International Conference on Pattern
  Recognition}, pages 0--0, 2020.

\bibitem{NNVALID2020FUHL}
Wolfgang Fuhl, Yao Rong, Thomas Motz, Michael Scheidt, Andreas Hartel, Andreas
  Koch, and Enkelejda Kasneci.
\newblock Explainable online validation of machine learning models for
  practical applications.
\newblock In {\em Proceedings of the International Conference on Pattern
  Recognition}, pages 0--0, 2020.

\bibitem{CAIP2019FuhlW}
W. Fuhl, W. Rosenstiel, and E. Kasneci.
\newblock 500,000 images closer to eyelid and pupil segmentation.
\newblock In {\em Computer Analysis of Images and Patterns, CAIP}, 11 2019.

\bibitem{WTDTE022017}
W. Fuhl, T. Santini, D. Geisler, T.~C. Kübler, and E. Kasneci.
\newblock Eyelad: Remote eye tracking image labeling tool.
\newblock In {\em 12th Joint Conference on Computer Vision, Imaging and
  Computer Graphics Theory and Applications (VISIGRAPP 2017)}, 02 2017.

\bibitem{WTDTWE092016}
W. Fuhl, T. Santini, D. Geisler, T.~C. Kübler, W. Rosenstiel, and E. Kasneci.
\newblock Eyes wide open? eyelid location and eye aperture estimation for
  pervasive eye tracking in real-world scenarios.
\newblock In {\em ACM International Joint Conference on Pervasive and
  Ubiquitous Computing: Adjunct publication -- PETMEI 2016}, 09 2016.

\bibitem{WTE032017}
W. Fuhl, T. Santini, and E. Kasneci.
\newblock Fast and robust eyelid outline and aperture detection in real-world
  scenarios.
\newblock In {\em IEEE Winter Conference on Applications of Computer Vision
  (WACV 2017)}, 03 2017.

\bibitem{CORR2017FuhlW1}
W. Fuhl, T. Santini, and E. Kasneci.
\newblock Fast camera focus estimation for gaze-based focus control.
\newblock In {\em CoRR}, 2017.

\bibitem{fuhl2018simarxiv}
W. Fuhl, T. Santini, T. Kuebler, N. Castner, W. Rosenstiel, and E. Kasneci.
\newblock Eye movement simulation and detector creation to reduce laborious
  parameter adjustments.
\newblock {\em arXiv preprint arXiv:1804.00970}, 2018.

\bibitem{WTCDAHKSE122016}
W. Fuhl, T. Santini, C. Reichert, D. Claus, A. Herkommer, H. Bahmani, K. Rifai,
  S. Wahl, and E. Kasneci.
\newblock Non-intrusive practitioner pupil detection for unmodified microscope
  oculars.
\newblock {\em Elsevier Computers in Biology and Medicine}, 79:36--44, 12 2016.

\bibitem{DWTE022017}
D. Geisler, W. Fuhl, T. Santini, and E. Kasneci.
\newblock Saliency sandbox: Bottom-up saliency framework.
\newblock In {\em 12th Joint Conference on Computer Vision, Imaging and
  Computer Graphics Theory and Applications (VISIGRAPP 2017)}, 02 2017.

\bibitem{glorot2010understanding}
Xavier Glorot and Yoshua Bengio.
\newblock Understanding the difficulty of training deep feedforward neural
  networks.
\newblock In {\em Proceedings of the thirteenth international conference on
  artificial intelligence and statistics}, pages 249--256, 2010.

\bibitem{he2016deep}
Kaiming He, Xiangyu Zhang, Shaoqing Ren, and Jian Sun.
\newblock Deep residual learning for image recognition.
\newblock In {\em Proceedings of the IEEE conference on computer vision and
  pattern recognition}, pages 770--778, 2016.

\bibitem{hinton2011transforming}
Geoffrey~E Hinton, Alex Krizhevsky, and Sida~D Wang.
\newblock Transforming auto-encoders.
\newblock In {\em International conference on artificial neural networks},
  pages 44--51. Springer, 2011.

\bibitem{howard2017mobilenets}
Andrew~G Howard, Menglong Zhu, Bo Chen, Dmitry Kalenichenko, Weijun Wang,
  Tobias Weyand, Marco Andreetto, and Hartwig Adam.
\newblock Mobilenets: Efficient convolutional neural networks for mobile vision
  applications.
\newblock {\em arXiv preprint arXiv:1704.04861}, 2017.

\bibitem{ioffe2015batch}
Sergey Ioffe and Christian Szegedy.
\newblock Batch normalization: Accelerating deep network training by reducing
  internal covariate shift.
\newblock {\em arXiv preprint arXiv:1502.03167}, 2015.

\bibitem{jaderberg2015spatial}
Max Jaderberg, Karen Simonyan, Andrew Zisserman, et~al.
\newblock Spatial transformer networks.
\newblock In {\em Advances in neural information processing systems}, pages
  2017--2025, 2015.

\bibitem{king2009dlib}
Davis~E King.
\newblock Dlib-ml: A machine learning toolkit.
\newblock {\em Journal of Machine Learning Research}, 10(Jul):1755--1758, 2009.

\bibitem{kingma2014adam}
Diederik~P Kingma and Jimmy Ba.
\newblock Adam: A method for stochastic optimization.
\newblock {\em arXiv preprint arXiv:1412.6980}, 2014.

\bibitem{krizhevsky2009learning}
Alex Krizhevsky, Geoffrey Hinton, et~al.
\newblock Learning multiple layers of features from tiny images.
\newblock 2009.

\bibitem{krizhevsky2012imagenet}
Alex Krizhevsky, Ilya Sutskever, and Geoffrey~E Hinton.
\newblock Imagenet classification with deep convolutional neural networks.
\newblock In {\em Advances in neural information processing systems}, pages
  1097--1105, 2012.

\bibitem{laptev2016ti}
Dmitry Laptev, Nikolay Savinov, Joachim~M Buhmann, and Marc Pollefeys.
\newblock Ti-pooling: transformation-invariant pooling for feature learning in
  convolutional neural networks.
\newblock In {\em Proceedings of the IEEE conference on computer vision and
  pattern recognition}, pages 289--297, 2016.

\bibitem{lecun1995convolutional}
Yann LeCun, Yoshua Bengio, et~al.
\newblock Convolutional networks for images, speech, and time series.
\newblock {\em The handbook of brain theory and neural networks},
  3361(10):1995, 1995.

\bibitem{li2017deep}
Rongjian Li, Tao Zeng, Hanchuan Peng, and Shuiwang Ji.
\newblock Deep learning segmentation of optical microscopy images improves 3-d
  neuron reconstruction.
\newblock {\em IEEE transactions on medical imaging}, 36(7):1533--1541, 2017.

\bibitem{long2015fully}
Jonathan Long, Evan Shelhamer, and Trevor Darrell.
\newblock Fully convolutional networks for semantic segmentation.
\newblock In {\em Proceedings of the IEEE conference on computer vision and
  pattern recognition}, pages 3431--3440, 2015.

\bibitem{lumini2019deep}
Alessandra Lumini and Loris Nanni.
\newblock Deep learning and transfer learning features for plankton
  classification.
\newblock {\em Ecological informatics}, 51:33--43, 2019.

\bibitem{madry2017towards}
Aleksander Madry, Aleksandar Makelov, Ludwig Schmidt, Dimitris Tsipras, and
  Adrian Vladu.
\newblock Towards deep learning models resistant to adversarial attacks.
\newblock {\em arXiv preprint arXiv:1706.06083}, 2017.

\bibitem{marcos2017rotation}
Diego Marcos, Michele Volpi, Nikos Komodakis, and Devis Tuia.
\newblock Rotation equivariant vector field networks.
\newblock In {\em Proceedings of the IEEE International Conference on Computer
  Vision}, pages 5048--5057, 2017.

\bibitem{1604}
Diego Marcos, Michele Volpi, and Devis Tuia.
\newblock Learning rotation invariant convolutional filters for texture
  classification.
\newblock In {\em 2016 23rd International Conference on Pattern Recognition
  (ICPR)}, pages 2012--2017. IEEE, 2016.

\bibitem{marin2019recipe1m}
Javier Marin, Aritro Biswas, Ferda Ofli, Nicholas Hynes, Amaia Salvador, Yusuf
  Aytar, Ingmar Weber, and Antonio Torralba.
\newblock Recipe1m+: A dataset for learning cross-modal embeddings for cooking
  recipes and food images.
\newblock {\em IEEE transactions on pattern analysis and machine intelligence},
  2019.

\bibitem{pinto2016curious}
Lerrel Pinto, Dhiraj Gandhi, Yuanfeng Han, Yong-Lae Park, and Abhinav Gupta.
\newblock The curious robot: Learning visual representations via physical
  interactions.
\newblock In {\em European Conference on Computer Vision}, pages 3--18.
  Springer, 2016.

\bibitem{qian1999momentum}
Ning Qian.
\newblock On the momentum term in gradient descent learning algorithms.
\newblock {\em Neural networks}, 12(1):145--151, 1999.

\bibitem{ICIP}
Rosemberg Rodriguez, Eva Dokladalova, and Petr Dokl{\'a}dal.
\newblock Rotation invariant cnn using scattering transform for image
  classification.
\newblock In {\em 2019 IEEE International Conference on Image Processing
  (ICIP)}, pages 654--658. IEEE, 2019.

\bibitem{ronneberger2015u}
Olaf Ronneberger, Philipp Fischer, and Thomas Brox.
\newblock U-net: Convolutional networks for biomedical image segmentation.
\newblock In {\em International Conference on Medical image computing and
  computer-assisted intervention}, pages 234--241. Springer, 2015.

\bibitem{RotCNN}
Rosemberg~Rodriguez Salas, Petr Dokl{\'a}dal, and Eva Dokladalova.
\newblock Red-nn: Rotation-equivariant deep neural network for classification
  and prediction of rotation.
\newblock 2019.

\bibitem{worrall2017harmonic}
Daniel~E Worrall, Stephan~J Garbin, Daniyar Turmukhambetov, and Gabriel~J
  Brostow.
\newblock Harmonic networks: Deep translation and rotation equivariance.
\newblock In {\em Proceedings of the IEEE Conference on Computer Vision and
  Pattern Recognition}, pages 5028--5037, 2017.

\bibitem{wu2018group}
Yuxin Wu and Kaiming He.
\newblock Group normalization.
\newblock In {\em Proceedings of the European conference on computer vision
  (ECCV)}, pages 3--19, 2018.

\bibitem{zhou2017oriented}
Yanzhao Zhou, Qixiang Ye, Qiang Qiu, and Jianbin Jiao.
\newblock Oriented response networks.
\newblock In {\em Proceedings of the IEEE Conference on Computer Vision and
  Pattern Recognition}, pages 519--528, 2017.

\end{thebibliography}
}

\end{document}